\pdfoutput=1

\documentclass[10pt,twocolumn,letterpaper]{article}

\usepackage{cvpr}              

\usepackage{graphicx}
\usepackage{amsmath}
\usepackage{amssymb}
\usepackage{booktabs}
\usepackage{tabularx}
\usepackage{bm}

\usepackage[pagebackref,colorlinks]{hyperref}

\usepackage[capitalize]{cleveref}
\crefname{section}{Sec.}{Secs.}
\Crefname{section}{Section}{Sections}
\Crefname{table}{Table}{Tables}
\crefname{table}{Tab.}{Tabs.}

\def\cvprPaperID{*****} 
\def\confName{CVPR}
\def\confYear{2023}

\begin{document}

\title{DialMAT: Dialogue-Enabled Transformer with \\Moment-Based Adversarial Training}

\author{
Kanta Kaneda\thanks{Equal contribution},
Ryosuke Korekata$^{*}$,
Yuiga Wada$^{*}$,
Shunya Nagashima$^{*}$,
Motonari Kambara,\\
Yui Iioka,
Haruka Matsuo,
Yuto Imai, 
Takayuki Nishimura,
and Komei Sugiura\\
Keio University\\
{\tt \small \{k.kaneda, rkorekata, yuiga, ng\_sh, motonari.k714, kmngrd1805,} \\ 
{\tt \small haruka.matsuo-25, ytim8812, t-nishimura, komei.sugiura\}@keio.jp}
}
\maketitle
\vspace{-3mm}
\begin{abstract}
\vspace{-4mm}
This paper focuses on the DialFRED task, which is the task of embodied instruction following in a setting where an agent can actively ask questions about the task.
To address this task, we propose DialMAT. DialMAT introduces Moment-based Adversarial Training, which incorporates adversarial perturbations into the latent space of language, image, and action. Additionally, it introduces a crossmodal parallel feature extraction mechanism that applies foundation models to both language and image. We evaluated our model using a dataset constructed from the DialFRED dataset and demonstrated superior performance compared to the baseline method in terms of success rate and path weighted success rate. The model secured the top position in the DialFRED Challenge, which took place at the CVPR 2023 Embodied AI workshop. 

\end{abstract}
\vspace{-3mm}

\vspace{-3mm}
\section{Introduction}
\vspace{-2mm}

In this paper, we focus on the task of embodied instruction following in a setting where an agent can ask questions to the human and utilize the information provided in the response to enhance its ability to complete the task effectively.


The main contributions of this paper are as follows:
\vspace{-2mm}
\begin{itemize}
    \setlength{\parskip}{0.2mm} %
    \setlength{\itemsep}{0.2mm} %
    \item We introduce Moment-based Adversarial Training (MAT) \cite{ishikawa-mat} to incorporate adversarial perturbations into the latent space of language, image, and action.
    \item We introduce a crossmodal parallel feature extraction mechanism to both language and image using foundation models \cite{radford2021learning,he2021debertav3}.
\end{itemize}
\vspace{-3mm}
\section{Problem Statement}
\vspace{-2mm}


This study focuses on the DialFRED task \cite{gao-dialfred}, which involves embodied instruction following. In this task, an agent has the ability to actively ask questions to the human user and utilize the information provided in the response to enhance its effectiveness in completing the task.
DialFRED is based on the standard benchmark ALFRED \cite{alfred20} for Vision-and-Language Navigation tasks involving object manipulation.
In the DialFRED setting, the robot can ask questions about the position of the object, its description, and the direction in which it should move.

The input and output for this task are defined as follows:
\vspace{-6mm}
\begin{itemize}
    \setlength{\parskip}{0.5mm} %
    \setlength{\itemsep}{0.2mm} %
    \item \textbf{Input:} Instructions for each subgoal, answers, and RGB images for each timestep.
    \item \textbf{Output:} Action taken at each timestep.
\end{itemize}
\vspace{-2mm}

\vspace{-3mm}
\section{Method}
\vspace{-2mm}
\begin{figure}[t]
    \centering
    \includegraphics[bb=0 0 3042 1808,height=50mm]{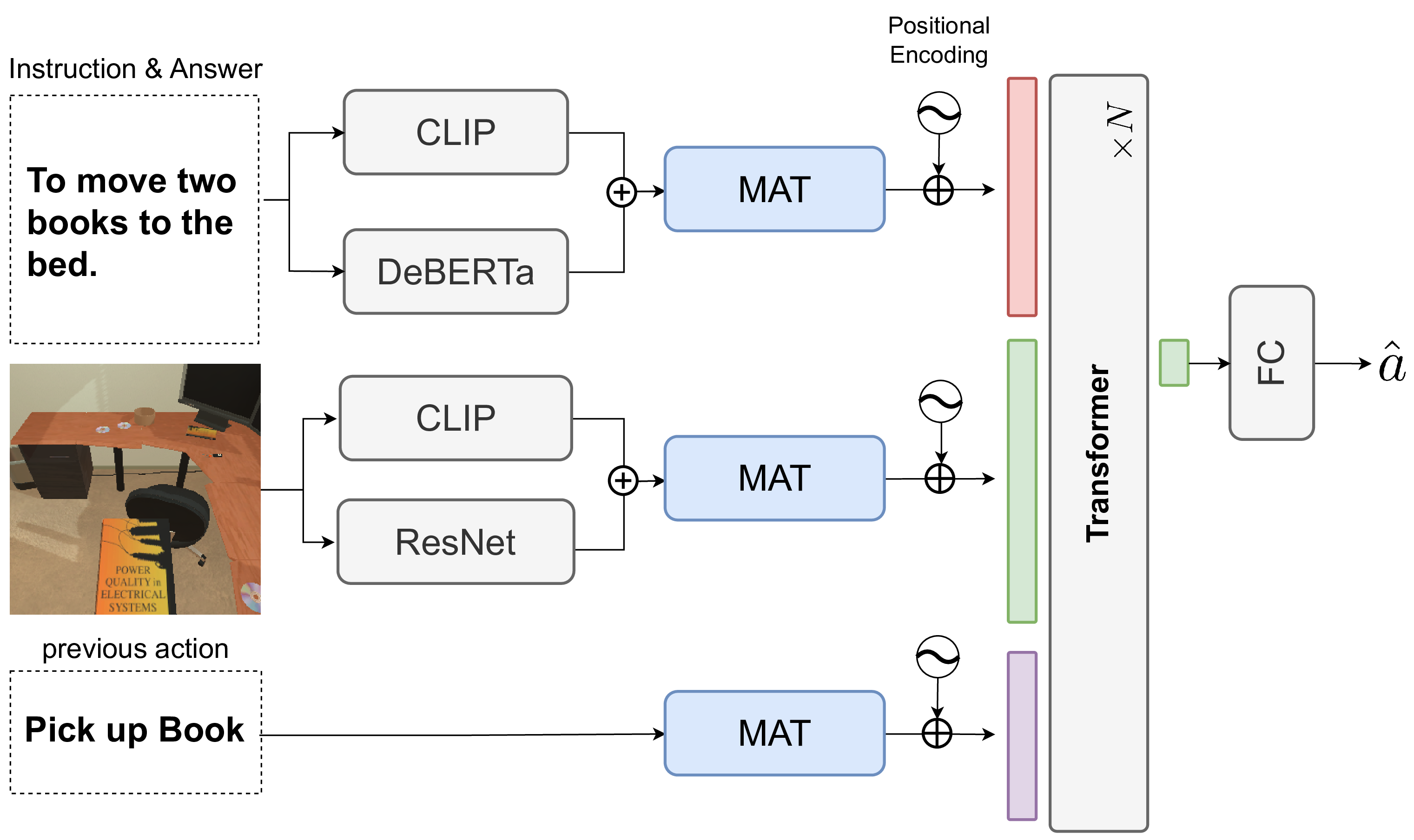}
    \vspace{-5mm}
    \caption{DialMAT consists of two main modules: Questioner and MAPer. Our method employs MAT to incorporate adversarial perturbations into the latent space of language, image, and action.}
    \label{fig:model}
    \vspace{-5mm}
\end{figure}
We propose DialMAT\footnote{\url{https://github.com/keio-smilab23/DialMAT}}, an extension of the Episodic Transformer \cite{pashevich2021episodic}. The proposed method consists of two main modules: Questioner and Moment-based Adversarial Performer (MAPer). Fig. \ref{fig:model} shows the overview of DialMAT.

The input $\bm{x}_t$ for time $t$ in our model is defined as $
\bm{x}_t = (D^{(k)}, l^{(k)}, v_t, \hat{a}_{t-1}), \; D^{(k)} = \{(q^{(k)}_i, s^{(k)}_i) \; | \; i = 0...N\}, 
$
where $l^{(k)}$, $v_t$ and $\hat{a}_{t-1}$ denote the instruction for the $k$th subgoal, perspective RGB image at time $t$, and previous action at time $t-1$, respectively.
Here, $\hat{a}$ is composed of a pair of action types and manipulated objects. Note that depending on the type of action, there may or may not be a manipulated object involved.
In addition, $D^{(k)}$ refers to the set of question-answer statements in the $k$th subgoal, which is composed of pairs of question statements $q^{(k)}_i$ and corresponding response statements $s^{(k)}_i$.

First, the Questioner determines which question needs to be asked at time $t$. This module has the same structure as the Questioner proposed in \cite{gao-dialfred}, and is composed of an LSTM with an attention mechanism.
The input in Questioner is $l^{(k)}$, and LSTM encoder and decoder perform multi-level classification. That is, at time $t$, it determines which of the Location, Appearance, and Direction questions to ask, and obtains the response $s^{(k)}$ for each question.

Second, the MAPer takes $\bm{x}_t$ as input and outputs the robot's behavior at time $t$.
First, it computes the respective embedded representations $h_\mathrm{ctxt}$ and $h_\mathrm{deb}$ from $l^{(k)}$ using CLIP \cite{radford2021learning} and DeBERTa v3 \cite{he2021debertav3}.
Then, following the MAT approach, we obtain the features $h_\mathrm{txt}$ by adding a learnable perturbation $\delta_\mathrm{txt}$ as $
    h_\mathrm{txt} = [h_\mathrm{ctxt}; \; h_\mathrm{deb}] + \delta_\mathrm{txt}
$.
Note that $\delta_\mathrm{txt}$ is updated based on the following steps. First, we compute the gradient of the loss function $E$ with respect to the perturbation $\nabla_{\delta}E$.
Next, we introduce two types of moving averages using $\nabla_{\delta}E$ as follows:
\vspace{-2mm}
\begin{align*}
    \bm{m}_t&=\rho_1\bm{m}_{t-1}+(1-\rho_1)\nabla_{\bm{\delta}}E(\bm{\delta}_t), \\
    \bm{v}_t&=\rho_2\bm{v}_{t-1}+(1-\rho_2)(\nabla_{\bm{\delta}}E(\bm{\delta}_t))^2
\vspace{-2mm}
\end{align*}
where $t$ denotes the current perturbation update step and $\rho_1, \rho_2$ denote the smoothing coefficients for each moving average.
Finally, using the above $\bm{m}_i$, $\bm{v}_i$, the update width of the perturbation $\Delta \delta_i$ is calculated as follows:
\vspace{-2mm}
\begin{gather*}
\hat{\bm{m}}_t=\frac{\bm{m}_t}{1-(\rho_1)^t}, \; \hat{\bm{v}}_t=\frac{\bm{v}_t}{1-(\rho_2)^t}, \; 
    \Delta\bm{\delta}_t=\eta\frac{\hat{\bm{m}}_t}{\sqrt{\hat{\bm{v}}_t+\varepsilon}}, \\ 
\bm{\delta}_{t+1}=\Pi_{\|\bm{\delta}\|\leq\epsilon}(\bm{\delta}_t+\frac{\Delta\bm{\delta}_t}{\|\Delta\bm{\delta}_t\|_F}),
\end{gather*}
where $\eta$, $\epsilon$, $\Pi_{\|\cdot\|\leq\epsilon}$ and $\|\cdot\|_F$ denote the learning rate of the MAT, minute value that prevents zero division, the projection onto the $\epsilon$-sphere and the Frobenius norm, respectively.
Next, we compute the respective embedded representations $h_\mathrm{cimg}$ and $h_\mathrm{res}$ from CLIP and ResNet \cite{he16resnet}. Then, we apply MAT to obtain the feature $h_\mathrm{img}$ by adding the perturbation $\delta_\mathrm{img}$ as $h_\mathrm{img} = [h_\mathrm{cimg}; \; h_\mathrm{res}] + \delta_\mathrm{img}$.    
Similarly, for $s^{(k)}$ and $a_{t-1}$, MAT is applied to obtain the respective latent representations $h_\mathrm{ans}$ and $h_\mathrm{act}$.
The following embedded representation $h^{1}$ is then input to the $N$-layer Transformer to obtain $h^{(N)}$:
\vspace{-2mm}
\begin{align*}
    h^{1} &= [h_\mathrm{txt}; \; h_\mathrm{ans}; \; h_\mathrm{img}; \; h_\mathrm{act}] + E_\mathrm{pos}, \\
    h^{i} &= \mathrm{Transformer}(h^{i-1}),
\end{align*}
where $E_{pos}$ refers to positional encoding and embeds the positional information of each token in $[h_\mathrm{txt}; \; h_\mathrm{ans}; \; h_\mathrm{img}; \; h_\mathrm{act}]$.
Finally, we obtain the predicted action $\hat{a}_t$  by applying the fully connected layer to $h^{(N)}$.
We define the loss function as the cross-entropy between $\hat{a}_t$ and expert action $a_t$.

\vspace{-2mm}
\section{Experimental Setup}
\vspace{-2mm}
In this study, we employ the same experimental settings as DialFRED\cite{gao-dialfred}.
In this study, we divided the DialFRED dataset as described in \cite{gao-dialfred}. 
As the test set of DialFRED is not publicly accessible, we further divided the valid\_unseen set to pseudo\_valid and pseudo\_test set, comprising 685 and 678 tasks, respectively.
We used the training set to train the MAPer and the valid\_seen set for reinforcement learning of the Questioner. 


\vspace{-2mm}
\section{Experimental Results}
\vspace{-2mm}

\begin{table}[t]
\centering
\caption{Quantitative comparison. The best scores are in bold.}
\normalsize
\vspace{-2mm}
\begin{tabular}{lccc}
\hline
{} & \multicolumn{2}{c}{Pseudo Test} & {Test} \\
{Method} & {SR$\uparrow$} & {PWSR$\uparrow$} & {SR$\uparrow$} \\ \hline
{Baseline\cite{gao-dialfred}} & {0.31} & {0.19} & {-} \\
{Ours (w/o MAT\cite{ishikawa-mat})} & {0.34} & {0.20} & {-} \\
{Ours (w/ CLIP text encoder\cite{radford2021learning})} & {0.35} & {0.22} & {-} \\
{Ours (MAT for action)} & {0.36} & {0.21} & {-} \\
{Ours (DialMAT)} & {\textbf{0.39}} & {\textbf{0.23}} & {\textbf{0.14}} \\
\hline
\end{tabular}
\label{tab:results}
\vspace{-4mm}
\end{table}
%
\begin{figure}[t]
    \centering
    \includegraphics[bb=0 0 182 88,height=30mm]{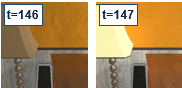}
    \vspace{-2mm}
    \caption{\small A successful subgoal prediction. In this case, the instruction was ``Move to the floor lamp, power on the floor lamp.''}
    \vspace{-6mm}
    \label{fig:qual}
\end{figure}
Table \ref{tab:results} shows the quantitative results of the baseline and the proposed methods. 
We evaluated the models by success rate (SR) and path weighted success rate (PWSR). 
Table \ref{tab:results} shows that the SR on the pseudo\_test set for the baseline and the proposed methods were $0.31$ and $0.39$, respectively. Therefore, the proposed method outperformed the baseline by $0.08$ points in SR.
Similarly, the proposed method also outperformed the baseline method in PWSR.

Fig.~\ref{fig:qual} shows the qualitative results. 
In this sample, the instruction was ``Move to the floor lamp, power on the floor lamp.'' In this case, the robot was required to move to the floor lamp and then turn it on. The proposed method was able to appropriately navigate to the floor lamp and successfully turn it on.
\vspace{-2mm}
\section{Conclusions}
\vspace{-2mm}

In summary, the contributions of this work are twofold:
\vspace{-2.5mm}
\begin{itemize}
    \setlength{\parskip}{0.2mm} %
    \setlength{\itemsep}{0.2mm} %
\item We introduced MAT to incorporate adversarial perturbations into the latent spaces.
\item We introduced crossmodal parallel feature extraction mechanisms to both language and image using foundation models.
\vspace{-2.5mm}
\end{itemize}
\textbf{Acknowledgements} This work was partially supported by JSPS KAKENHI Grant Number JP23H03478, JST Moonshot, and NEDO.

{\small
\bibliographystyle{ieee_fullname}
\bibliography{reference}
}

\end{document}